\documentclass[10pt,twocolumn,letterpaper]{article}

\usepackage[pagenumbers]{cvpr} 

\usepackage{multirow}
\usepackage{algorithm}
\usepackage{algorithmic}
\usepackage{amsmath}
\usepackage{wrapfig}
\definecolor{cvprblue}{rgb}{0.21,0.49,0.74}
\usepackage[pagebackref,breaklinks,colorlinks,allcolors=cvprblue]{hyperref}

\title{FoPru: Focal Pruning for Efficient Large Vision-Language Models}

\author{
\textbf{Lei Jiang\textsuperscript{1}}\quad
\textbf{Weizhe Huang\textsuperscript{1}}\quad
\textbf{Tongxuan Liu\textsuperscript{1,*}} \quad
\textbf{Yuting Zeng\textsuperscript{1} }\quad
\textbf{Jing Li\textsuperscript{1} } \\
\textbf{Lechao Cheng\textsuperscript{2,*}} \quad
\textbf{Xiaohua Xu\textsuperscript{1,*}}\\
\textsuperscript{1}University of Science and Technology of China \quad
\textsuperscript{2}Hefei University of Technology\\
\tt\small{\{jianglei0510, hwz871982879, tongxuan.ltx, yuting\_zeng\}@mail.ustc.edu.cn}\\
\tt\small{\{lj, xiaohuaxu\}@ustc.edu.cn} 
\tt\small{chenglc@hfut.edu.cn}\\
}

\begin{document}
\maketitle
\begin{abstract}
Large Vision-Language Models (LVLMs) represent a significant advancement toward achieving superior multimodal capabilities by enabling powerful Large Language Models (LLMs) to understand visual input. 
Typically, LVLMs utilize visual encoders, such as CLIP, to transform images into visual tokens, which are then aligned with textual tokens through projection layers before being input into the LLM for inference. 
Although existing LVLMs have achieved significant success, their inference efficiency is still limited by the substantial number of visual tokens and the potential redundancy among them.
To mitigate this issue, we propose \textbf{Fo}cal \textbf{Pru}ning (FoPru), a training-free method that prunes visual tokens based on the attention-based token significance derived from the vision encoder. 
Specifically, we introduce two alternative pruning strategies: 1) the rank strategy, which leverages all token significance scores to retain more critical tokens in a global view; 2) the row strategy, which focuses on preserving continuous key information in images from a local perspective. 
Finally, the selected tokens are reordered to maintain their original positional relationships. Extensive experiments across various LVLMs and multimodal datasets demonstrate that our method can prune a large number of redundant tokens while maintaining high accuracy, leading to significant improvements in inference efficiency.

\end{abstract}    
\def\thefootnote{*}\footnotetext{Corresponding authors.}
\section{Introduction}
\label{sec:introduction}

In recent years, Large Vision Language Models (LVLMs)~\cite{li2023blip, zhu2023minigpt, liu2024visual, bai2023qwen, team2023gemini} have exhibited remarkable capabilities in diverse multimodal scenarios, propelling advancements in intricate tasks such as image and language comprehension. These models typically involve a substantial number of visual tokens, often ranging from hundreds to thousands~\cite{cai2024vip}. The large quantity of visual tokens significantly amplifies the training and inference costs of LVLMs~\cite{chen2024image}. 

To alleviate the issues of excessive visual tokens in LVLMs, researchers have proposed a series of visual token compression methods~\cite{bolya2022token, chen2024image, shang2024llava}. For instance, Q-Former~\cite{li2023blip} and Resampler~\cite{bai2023qwen} utilize cross-attention and a set of learnable queries to extract the most relevant visual tokens and manage their quantity. Abstractor~\cite{cha2024honeybee} and LDP~\cite{chu2023mobilevlm, chu2024mobilevlm} employ convolutional layers to aggregate visual features, thereby generating compressed visual tokens. DenseConnector~\cite{yao2024dense} regulates the number of visual tokens through learnable MLP layers. In DocKylin~\cite{zhang2024dockylin}, redundant regions in images are identified and removed using image gradient information, while a k-means clustering method is employed to extract relevant tokens from a vast pool of visual tokens. However, these compression methods generally require retraining LVLMs, rendering them unsuitable for direct application to pre-existing general-purpose LVLMs. 

\begin{figure}[t]
\centering
\includegraphics[width=\columnwidth]{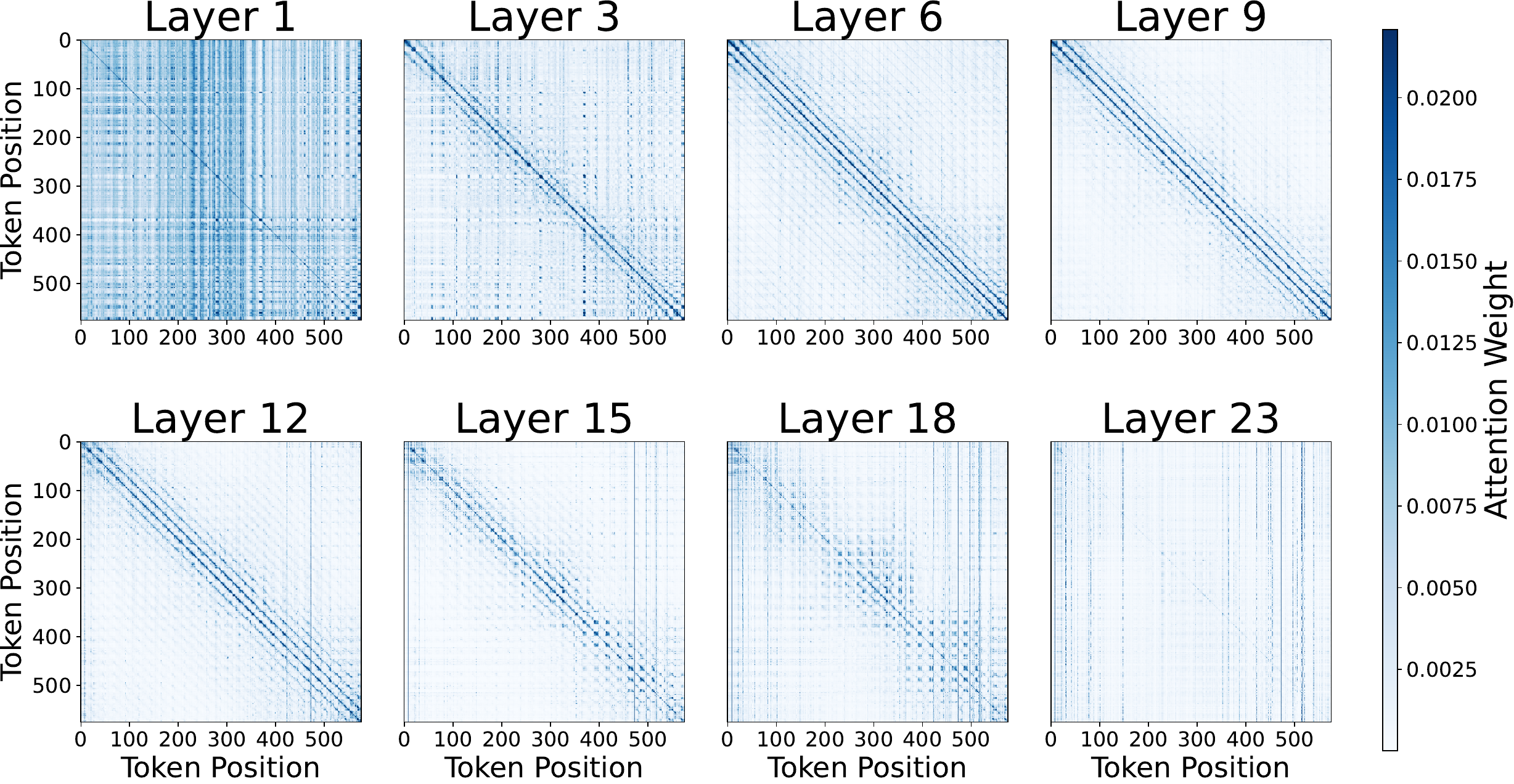} 
\caption{The attention map of CLIP in different layers.}
\label{fig:attention}
\vspace{-5mm}
\end{figure}

We observe that the deep layers in the visual encoder exhibit an imbalance in attention distribution, where attention is concentrated on a limited number of tokens, as shown in Figure~\ref{fig:attention}. This suggests that during the visual encoding stage, a small subset of visual tokens already captures critical visual information, while a significant proportion of tokens are likely unimportant and redundant. Motivated by this observation, we propose \textbf{Fo}cal \textbf{Pru}ning (FoPru), a training-free token pruning approach that can be seamlessly applied to various LVLMs. 
Specifically, FoPru consists of three stages. First, the Token Significance stage leverages attention scores derived from the visual encoder to calculate the significance of each token. The Token Pruning stage then prunes visual tokens based on these significance scores. In the Token Reordering stage, tokens are reordered according to their original positions, maintaining their relative positional relationships. Within the Token Pruning stage, we further introduce two alternative pruning strategies: Rank Pruning, which retains the most critical tokens from a global perspective, and Row Pruning, which preserves local continuous tokens row by row.

To validate the effectiveness of FoPru, we conduct experiments on diverse models and datasets. 
The results demonstrate that the FoPru approach significantly reduces visual tokens while achieving remarkable performance across multiple datasets. 
\textcolor{black}{Remarkably, even at an extreme token retention ratio of 0.2\% (retaining as few as 5 tokens), FoPru maintains approximately 60\% accuracy on the Ai2D\cite{kembhavi2016diagram} and SQA\cite{lu2022learn} datasets.}
Additionally, using only 25\% of visual tokens FoPru yields accuracy within a 1\% margin on 
MMMU \cite{yue2024mmmu}, SQA, and POPE \cite{li2023evaluating} datasets, with a maximum improvement of 2.52X in Time To First Token (TTFT) and 1.24X in Time Per Output Token (TPOP), thereby substantially enhancing inference efficiency for LVLMs.

\begin{figure*}[ht]
\centering
\includegraphics[width=\textwidth]{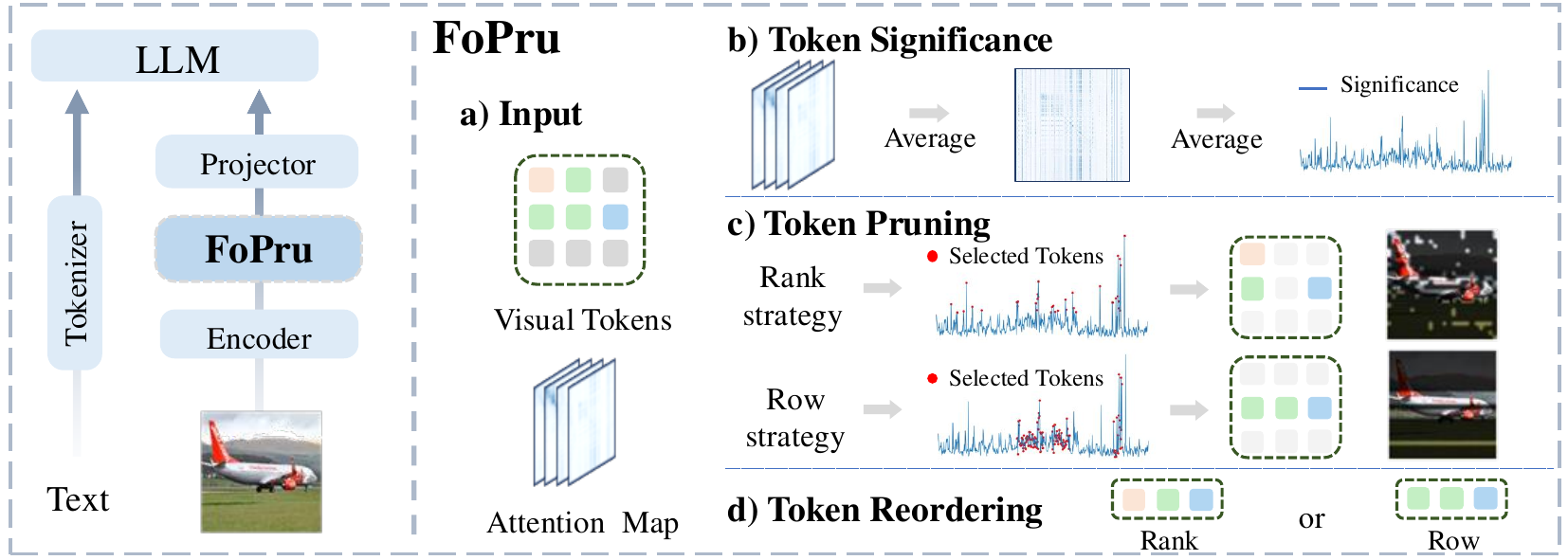}
\caption{The framework of Focal Pruning for LVLMs. First, we obtain the attention map in the vision encoder and calculate the token significance scores based on it. Next, we utilize alternative pruning strategies to prune the less important tokens and finally reorder the remaining tokens to recover relative positions.}
\label{fig:method}
\vspace{-5mm}
\end{figure*}

The core contributions of this paper are as follows:
\begin{enumerate}
    \item \textbf{Proposing a General Visual Token Pruning Method}: We introduce \textbf{F}ocal \textbf{P}runing (FoPru), a training-free approach that achieves pruning based on the attention distribution provided by the visual encoder itself, which is applicable to various LVLMs. 
    \item \textbf{Developing a Framework Supporting Multiple Token Pruning Strategies}: We construct a framework that implements various pruning strategies based on the distinct characteristics of images.
    \item \textbf{Validating the Effectiveness of the Method Across Multiple Datasets and Models}: Extensive experiments on diverse benchmark datasets and LVLMs demonstrate that our FoPru can significantly reduce the number of visual tokens and achieve efficient inference while maintaining accuracy.
\end{enumerate}

\section{Related Work}
\label{sec:related-work}

\subsection{Large Vision-Language Models (LVLMs)}
Large Language Models (LLMs), such as GPT-4 \cite{achiam2023gpt} and Llama \cite{touvron2023llama}, have achieved remarkable progress and demonstrated excellent capabilities in a wide range of natural language understanding tasks. In light of the great advantages of LLMs, recent Large Vision-Language Models (LVLMs) \cite{yin2023survey} transform image information into visual tokens and align them with textual tokens as inputs to the LLMs, resulting in significant advancements in multimodal capabilities. First, BLIP-2 \cite{li2023blip} is a pioneering model that employs a learnable and lightweight Q-Former to bridge a vision encoder and a LLM. This model freezes the two components separately and performs two-stage pre-training, thereby achieving cross-modal alignment. Through improving BLIP-2, many researchers produce various excellent open-source LVLMs \cite{zhu2023minigpt, bai2023qwenvl, wang2023cogvlm}. 
MiniGPT-4 \cite{zhu2023minigpt} simplifies alignment between a visual encoder and the LLM by using a single linear projector.
LLava \cite{liu2024visual,liu2024llava15} collects instruction data generated by GPT4 to fine-tune both the LLM and the visual-to-text projection matrix through end-to-end visual instruction tuning. There are also emerging powerful proprietary models, such as GPT-4V \cite{yang2023dawn}, Qwen-VL-Max \cite{bai2023qwenvl}, and Gemini \cite{reid2024gemini}, which show top-tier multimodal capabilities across a variety of visual-language tasks.

\subsection{Token Reduction in LVLMs}
Although current LVLMs have achieved remarkable vision-language understanding abilities, concerns remain that their inference efficiency is limited by their auto-regressive generation paradigm and potential token redundancy, especially when dealing with a large number of visual and textual tokens. In the literature, some researchers have attempted to employ various methods to mitigate this issue.
Q-Former \cite{li2023blip} and Resampler \cite{bai2023qwen} utilize cross-attention and a set of learnable queries to obtain the most relevant tokens to control their quantity. Abstractor \cite{cha2024honeybee} and LDP \cite{chu2023mobilevlm, chu2024mobilevlm} employ convolutional layers to aggregate visual features, generating compressed tokens. DenseConnector controls token quantity through learnable MLP layers. DocKylin \cite{zhang2024dockylin} leverages Adaptive Pixel Slimming (APS) and Dynamic Token Slimming (DTS) to compress visual content at the pixel and token levels. 
TokenPacker \cite{li2024tokenpacker} proposes a novel visual projector, which injects enriched high-resolution characteristics into a coarse low-resolution one to generate the condensed visual tokens. The aforementioned training-based methods require substantial resources to train specific LVLMs and lack generality. 
Instead, FastV \cite{chen2024image} is a recent training-free method that performs layer-level pruning to discard tokens with low attention scores in the LLM backbone. 
Similarly, PruMerge and PruMerge+ \cite{shang2024llava} are training-free methods that cluster visual tokens, updating key tokens through k-nearest neighbor weighted averaging.
However, these approaches overlook the importance of encoder's internal attention map and the mode collapse we have discovered in its deeper layers, thereby limiting their ability to guide token redundancy identification and leading to suboptimal pruning performance.

\begin{figure}[t]
\centering
\includegraphics[width=0.95\columnwidth]{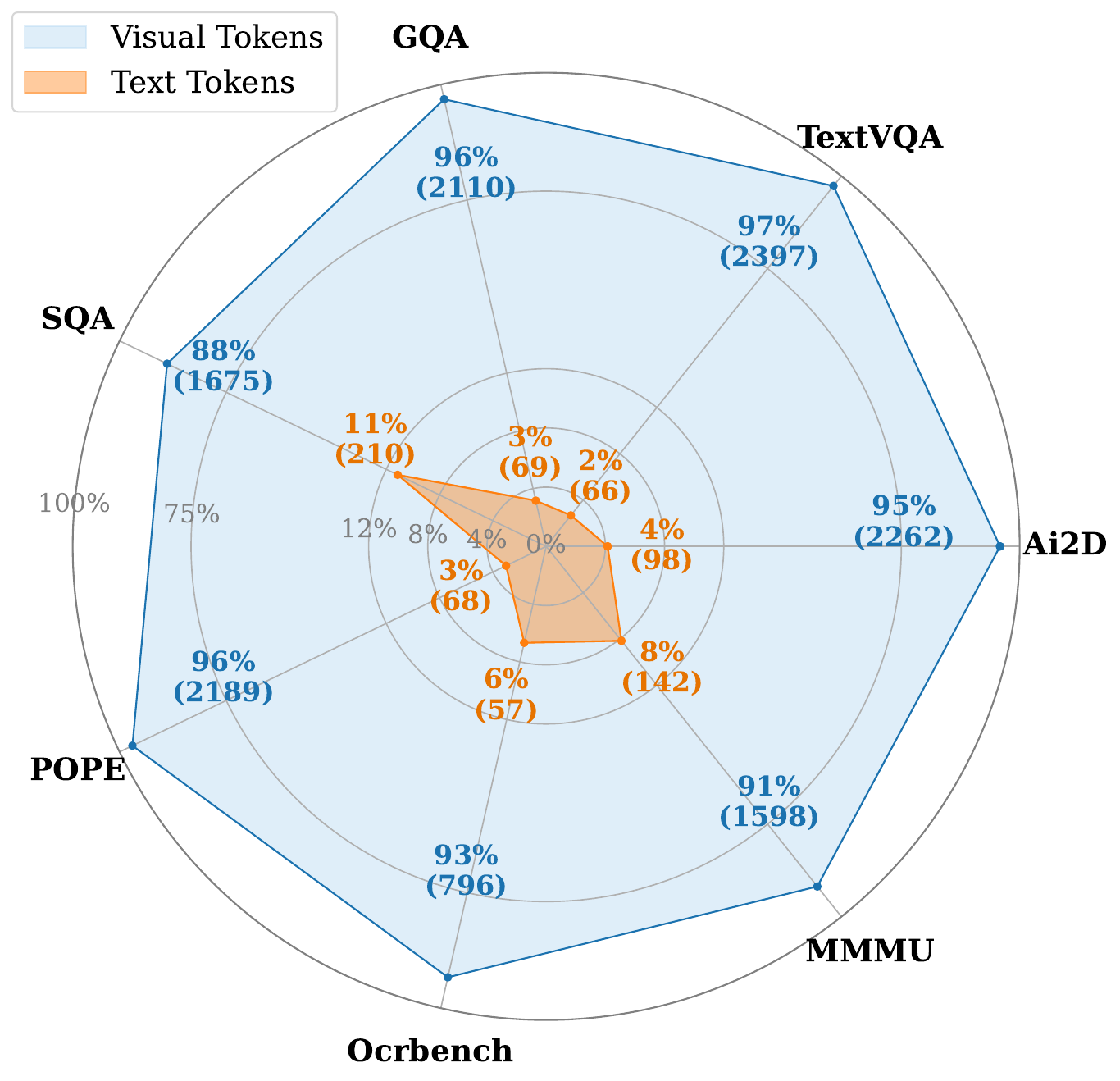}
\caption{The proportion of visual tokens and textual tokens in seven different datasets.}
\label{fig:visual_text}
\vspace{-5mm}
\end{figure}

\section{Preliminary}
\label{sec:preliminary}


LVLMs are aimed at generating textual responses based on input images and instructions \cite{yin2023survey}. 
A typical LVLM consists of three key modules:
a vision encoder, an advanced LLM, and a projector, which serves as a bridge for modality alignment. First, the vision encoder transforms the input image into visual embeddings $\mathbf{E_v}$, often utilizing the ViT architecture \cite{dosovitskiy2020vit}. Next, the projector converts these visual embeddings into visual tokens $\mathbf{T_v}$ by mapping them into the text space, making them understandable to the LLM. Given the generated visual tokens $\mathbf{T_v}$ and instructions' textual tokens $\mathbf{T_t}$, the LLM then produces the $L$-length output response $\mathbf{Y}$ in an auto-regressive manner based on the following probability distribution:
\begin{equation}
    P(\mathbf{Y}|\mathbf{T_t}, \mathbf{T_v}) = \prod_{i=1}^L P(\mathbf{Y}_i|\mathbf{T_t}, \mathbf{T_v}, \mathbf{Y}_{<i}). 
\end{equation}
As shown in the formula, the inference efficiency and memory requirements of LVLMs heavily depend on the length of the input tokens that the LLM needs process, which consist of both textual and visual tokens. In fact, due to the auto-regressive nature of LLM decoding, the computational complexity of the LLM is proportional to the square of the input token length. This indicates that reducing the input tokens is crucial for improving the inference efficiency of LVLMs.

\subsection{Token Redundancy Analysis}
\label{sec:token_redundancy}

In this subsection, we present important data analysis on the redundancy of visual tokens in LVLMs. First, we analyze the high proportion of visual tokens among the input tokens to the LLM. Then, we observe the imbalanced attention distribution in the vision encoder, which indicates the presence of numerous unimportant and redundant tokens in its output.

\paragraph{\textit{High Proportion of Visual Tokens.}}
We randomly select 10 samples from each of seven multimodal datasets and count the number and proportion of visual tokens and textual tokens using \texttt{LLaVA-NeXT-8B}~\cite{li2024llavanext-strong}. The average results are presented in Figure \ref{fig:visual_text}. This statistic suggests that visual tokens dominate the input tokens to the LLM, which aligns with findings from other research on other LVLMs~\cite{chen2024image}. 
This high proportion of visual tokens affects inference efficiency, suggesting that some visual tokens might be unimportant and could  be pruned to improve processing speed.

\paragraph{\textit{Imbalance Attention in Vision Encoder.}}
To further explore the redundancy of visual tokens, we take a step back to investigate the preceding visual encoder, from which the visual tokens originate. Inspired by \cite{he2023mitigating}, we quantify and visualize the attention maps from selected layers (Layer 1 to 23) in the CLIP model, as shown in Figure~\ref{fig:attention}. We observe that while the shallow layers exhibit relatively balanced attention distribution, the deep layers present a phenomenon known as mode collapse, where over 80\% of the attention is concentrated on less than 25\% of the tokens. This imbalance in attention suggests that only a few visual tokens with high attention scores contain critical visual information.

\section{Methodology}
\label{sec:methodology}

\subsection{Overview}

Figure \ref{fig:method} illustrates the overall architecture of FoPru for LVLMs. 
FoPru identifies and prunes redundant visual tokens before they reach the LLM.
First, we leverage the attention maps from the vision encoder to calculate the significance score of each token in the Token Significance stage. 
Then we introduce two Token Pruning Strategies: rank and row strategy, which focus on retaining global key visual information and local continuous information, respectively.
Next, in the Token Reordering stage, the selected tokens are reordered to restore their relative positional information.
An overview of this algorithm is provided in Algorithm \ref{alg:token_pruning}.

\begin{figure}[t!]
\centering
\includegraphics[width=\columnwidth]{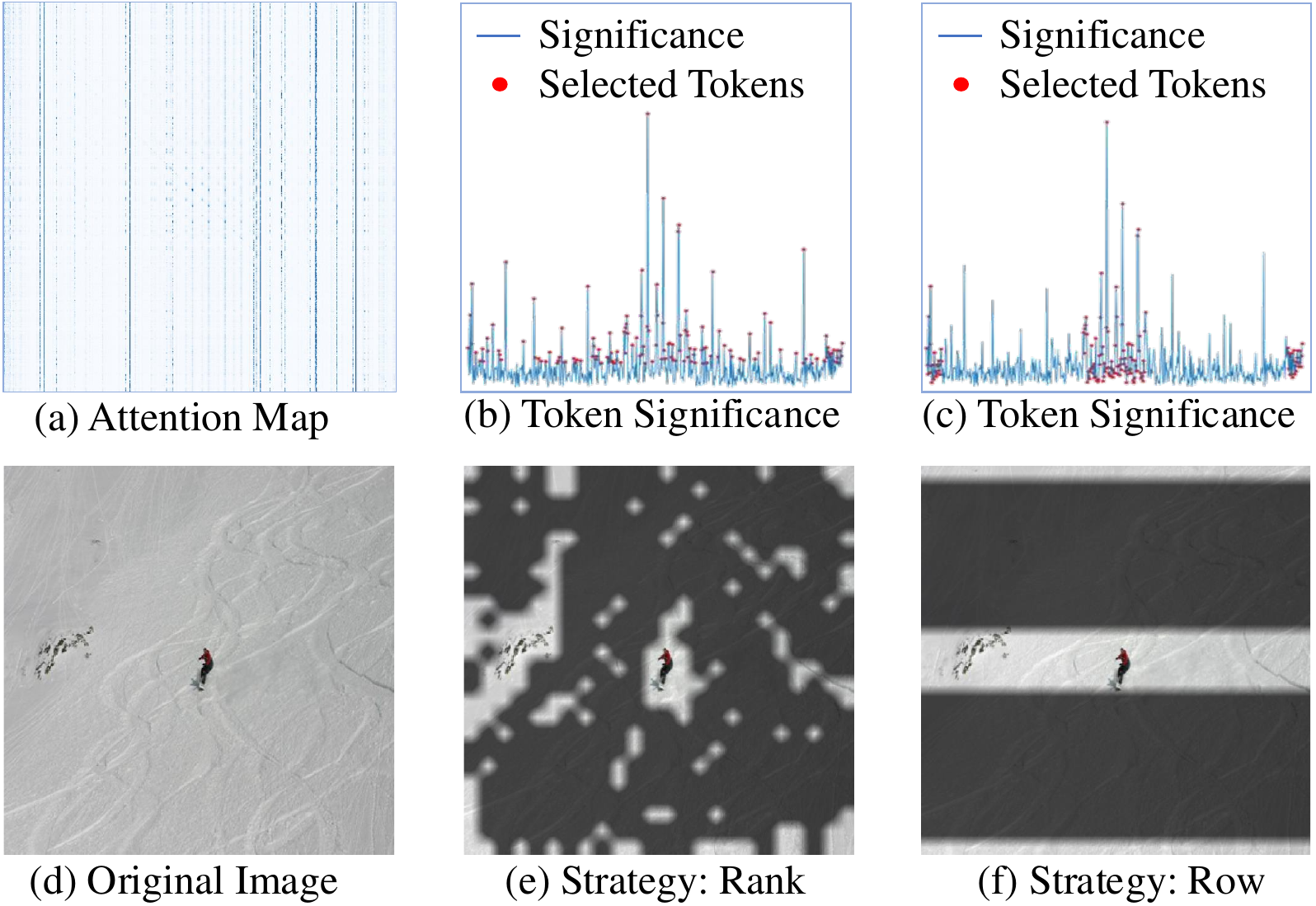} 
\caption{
The CLIP model processes the input image (d) to generate the attention map (a), on which the token significance score is computed in FoPru. Rank and row pruning strategies are then applied, shown in (b) and (c), respectively. Figures (e) and (f) highlight the image regions selected by the rank and row strategies.
}
\vspace{-3mm}
\label{fig:example}
\end{figure}

\subsection{Token Significance}

To prune redundant tokens for accelerating inference, it is crucial to identify the significance scores for each token.
The pruning process is guided by multi-head attention map $\mathbf{A}_k \in \mathbb{R}^{N\times N}$ for $k \in \{1, \dots, H\}$, extracted from the penultimate layer of the encoder, captures spatial dependencies among tokens. Here, $N$ represents the number of visual tokens,
and $H$ denotes the number of attention heads.
This layer is chosen because, in LLaVA, its image feature output provides the primary visual representation, which is subsequently aligned with text by projecting it into the visual token space.
First, we average the attention weights $\mathbf{A}_{k}$ across all heads as follows:
\begin{equation}
    \mathbf{\overline{A}} = \frac{1}{H} \sum_{k=1}^{H} \mathbf{A}_k.
\end{equation}
Next, we take the average along the last two dimensions of $\mathbf{\overline{A}}$ to get the average attention score for each dimension:
\begin{equation}
    \mathbf{s}_{\text{1}} = \frac{1}{N} \sum_{j=1}^{N} \mathbf{\overline{A}}[:, j], \
    \mathbf{s}_{\text{2}} = \frac{1}{N} \sum_{i=1}^{N} \mathbf{\overline{A}}[i, :],
\end{equation}
where $\mathbf{s}_{1}$ represents the average attention across columns, and $\mathbf{s}_{2}$ represents the average across rows.
Then, we compute the variance of these two vectors, $\text{Var}_{1}$ and $\text{Var}_{2}$, to identify the direction with greater data dispersion:
\(
    \text{Var}_{\text{1}} = \text{Var}(\mathbf{s}_{1}),
    \text{Var}_{\text{2}} = \text{Var}(\mathbf{s}_{2}).
\)
Intuitively, the dimension with higher variance has a more dispersed data distribution, making important tokens stand out more prominently. We therefore select the vector with the larger variance as the final token significance score $\mathbf{Sig}$:
\begin{equation}
    \mathbf{Sig} = 
    \begin{cases} 
    \mathbf{s}_{\text{1}}, & \text{if } \ \text{Var}_{\text{1}} > \text{Var}_{\text{2}} \\ 
    \mathbf{s}_{\text{2}}, & \text{otherwise}
    \end{cases}.
    \label{equ:sig}
\end{equation}

\noindent
\subsection{Token Pruning} 

We propose two alternative strategies to implement token pruning in FoPru. We denote $r$ as the token retention ratio, a predefined hyperparameter.

\noindent
\textbf{\textit{Rank strategy}}
To capture the core global visual information, we first calculate each visual token's significance score $\mathbf{Sig} \in \mathbb{R}^{1\times N}$ using Eq.~\ref{equ:sig}. Visual tokens are then globally ranked by these significance scores, and only the top \( N \times r\% \) tokens are retained while the less significant tokens are discarded.
As shown in Figure~\ref{fig:example}b, the rank strategy selects key tokens, with Figure~\ref{fig:example}e highlighting their corresponding image locations, effectively maintaining essential semantic information.
This rank strategy enables the model to concentrate on the most globally informative visual features, reducing potential interference from redundant tokens.

\noindent
\textbf{\textit{Row Strategy}}
To implement a structured token pruning mechanism, we begin by reshaping the one-dimensional visual token significance score vector $\mathbf{Sig}$, into a two-dimensional layout $\mathbf{Sig}_{\text{grid}} \in \mathbb{R}^{n\times n}$ that reflects the relative spatial positions within the image, where \(N = n \times n\). 
Considering that textual information is predominantly organized horizontally, we compute an aggregate significance score for each of the \( n \) rows.
Based on their accumulated significance scores, we select \( n \times r\% \) rows, denoted as $\mathbf{Sig}_{\text{grid}}[\mathcal{R}, :]$, and reshape them into a one-dimensional sequence. Here $\mathcal{R}$ is the index set of the most significant rows, as shown in Figure~\ref{fig:example}c and f.
The row strategy approach preserves spatial coherence and essential horizontal features, maintaining the structural integrity of the original image.

\subsection{Token Reordering} 
To preserve the relative positional relationships among the selected visual tokens, a reordering process is applied. 
The indices of selected tokens, determined by the rank or row strategy, are denoted as $\mathcal{I}_{\text{idx}}$ with shape \( (1, N \times r\%) \). 
Sorting $\mathcal{I}_{\text{idx}}$ in ascending order yields $\mathcal{I}_{\text{sorted}}$, which maintains the tokens' original spatial positions. 
Using $\mathcal{I}_{\text{sorted}}$, we obtain the pruned and spatially ordered set of visual tokens, $\mathbf{T}_{\text{pruned}} = \{ t_i \mid i \in \mathcal{I}_{\text{sorted}} \}$.
This reordering is crucial for preserving the contextual integrity and continuity necessary for accurate inference by LVLMs.

After reordering, the pruned visual tokens are first projected to align with the textual tokens in terms of modality. 
Then the visual tokens are combined with the textual tokens and jointly input into the LLM. 
This integrated input enables the LLM to utilize only the most relevant visual information to perform inference. 
As a result, the LVLMs can achieve greater computational efficiency while maintaining or even enhancing the final accuracy performance.

\begin{algorithm}
\caption{FoPru: Visual Token Pruning for LVLMs}
\label{alg:token_pruning}
\begin{algorithmic}[1]

\REQUIRE Visual tokens $\mathbf{T} = \{ t_1, t_2, \dots, t_N \}$, $N$ is the number of tokens,
attention maps $\mathbf{A}_k \in \mathbb{R}^{N\times N}$ from CLIP's penultimate layer for each attention head $k \in \{1, \dots, H\}$,
ratio $r$, strategy $s \in \{\textit{Rank}, \textit{Row}\}$.
\ENSURE Pruned visual tokens $\mathbf{T}_{\text{pruned}}$ for LVLM inference.

\STATE \textbf{Step 1: Token Significance}
    \STATE $\overline{\mathbf{A}} = \frac{1}{H} \sum_{k=1}^{H} \mathbf{A}_k$ 
    \STATE $\mathbf{s}_{1} = \frac{1}{N} \sum_{j=1}^{N} \overline{\mathbf{A}}[:, j]$, \quad $\mathbf{s}_{2} = \frac{1}{N} \sum_{i=1}^{N} \overline{\mathbf{A}}[i, :]$
    \STATE $\mathbf{Sig} = \mathbf{s}_{1}$ if $\operatorname{Var}(\mathbf{s}_{1}) > \operatorname{Var}(\mathbf{s}_{2})$ else $\mathbf{s}_{2}$

\STATE \textbf{Step 2: Token Selection}
    \IF{$s = \text{Rank}$}
       \STATE $\mathcal{I}_{idx} = \operatorname{sort\_{idx}}(\mathbf{Sig},desc)[:N \times r\%]$
    \ELSIF{$s = \text{Row}$}
        \STATE $\mathbf{Sig}_{\text{grid}} = \operatorname{reshape}(\mathbf{Sig}, (\sqrt{N}, \sqrt{N}))$ 
        \STATE $\mathcal{R} = \operatorname{sort\_{idx}}(\operatorname{sum\_row}(\mathbf{Sig}_{\text{grid}}),desc)[:\sqrt{N} \times r\%]$
        \quad // Returns sorted indices in descending order
        \STATE $\mathcal{I}_{idx} = \operatorname{idx}(\operatorname{flatten}(\mathbf{Sig}_{\text{grid}}[\mathcal{R}, :]$))
    \ENDIF

\STATE \textbf{Step 3: Token Reordering}
    \STATE $\mathcal{I}_{\text{sorted}} = \operatorname{sort\_value}(\mathcal{I}_{idx}, asc)$ 
    \quad // Sorts selected indices in ascending order for spatial coherence
    \STATE $\mathbf{T}_{\text{pruned}} = \{ t_i \mid i \in \mathcal{I}_{\text{sorted}} \}$ 
\end{algorithmic}
\end{algorithm}

\section{Experiments}

\begin{table*}[ht]
\centering
\setlength\tabcolsep{3.8 pt}
\renewcommand{\arraystretch}{0.8}
\begin{tabular}{cc|ccccccc|ccccc} 
\toprule

\multirow{3}{*}{\textbf{Model}} & \multirow{3}{*}{\textbf{Ratio}} & \multicolumn{7}{c|}{\textbf{Accuracy Performance}} & \multicolumn{5}{c}{\textbf{Inference Efficiency}} \\ 
 &  & \multirow{2}{*}{\textbf{Ai2D}} & \multirow{2}{*}{\textbf{GQA}} & \multirow{2}{*}{\textbf{MMMU}} & \multirow{2}{*}{\textbf{SQA}} & \multirow{2}{*}{\textbf{POPE}} 
 & \multirow{2}{*}{\textbf{TextVQA}}
 & \multirow{2}{*}{\textbf{Ocrbench}} & \multicolumn{2}{c}{\textbf{TTFT}} & \multicolumn{2}{c}{\textbf{TPOT}} & \textbf{GPU} \\ 
 &  &  &  &  &  &  &  &  & (ms & $S_p$)  & (ms/tok. & $S_p$)  & GB \\ 

\midrule
\multirow{4}{*}{\begin{tabular}[c]{@{}c@{}}LLaVA-\\NeXT-8B\end{tabular}} & 100\% & 71.66 & 65.38 & 40.22 & 79.44 & 87.84 & 65.43 & 54.90 & 94 & - & 25.61 & - & 17.88  \\ 
\cmidrule{3-14}
 & 75\% & 70.69 & 65.21 & 39.78 & \textbf{79.91} & \textbf{87.87} & 64.14 & 53.20 & 88 & 1.18x & 25.40 & 1.01x & 17.33  \\ 
 & 50\% & 70.02 & 64.82 & 39.67 & 79.39 & 87.13 & 62.86 & 49.50 & 57 & 1.66x & 24.80 & 1.03x & 16.98  \\ 
 & 25\% & 68.01 & 63.00 & 39.22 & 79.27 & 86.88 & 61.24 & 45.90 & 52 & \textbf{1.83x} & 24.05 & \textbf{1.07x} & 16.98  \\ 
\midrule
\multirow{4}{*}{\begin{tabular}[c]{@{}c@{}}LLaVA-\\1.6-7B\end{tabular}} & 100\% & 66.58 & 64.24 & 35.10 & 73.21 & 87.61 & 64.90 & 52.20 & 88 & - & 23.70 & - & 16.15  \\ 
\cmidrule{3-14}
 & 75\% & 65.54 & 64.13 & \textbf{37.00} & 73.19 & \textbf{87.93} & 63.00 & 51.30 & 81 & 1.09x & 23.55 & 1.01x & 15.44  \\ 
 & 50\% & 64.83 & 63.83 & \textbf{37.33} & 72.91 & \textbf{87.93} & 63.01 & 47.70 & 60 & 1.47x & 23.05 & 1.03x & 14.81  \\ 
 & 25\% & 64.35 & 62.26 & \textbf{36.67} & 72.41 & 86.83 & 60.81 & 44.60 & 50 & \textbf{1.78x} & 21.97 & \textbf{1.08x} & 14.57  \\ 
\midrule
\multirow{4}{*}{\begin{tabular}[c]{@{}c@{}}LLaVA-\\1.6-13B\end{tabular}} & 100\% & 70.30 & 65.37 & 35.90 & 75.85 & 87.56 & 67.10 & 55.10 & 198 & - & 38.30 & - & 29.53  \\ 
\cmidrule{3-14}
 & 75\% & 69.56 & \textbf{65.43} & \textbf{36.44} & \textbf{75.95} & \textbf{87.78} & 66.03 & 53.90 & 153 & 1.29x & 36.06 & 1.06x & 28.53  \\ 
 & 50\% & 68.98 & 65.15 & \textbf{37.11} & \textbf{76.23} & \textbf{87.84} & 64.33 & 50.10 & 116 & 1.70x & 33.35 & 1.15x & 27.53   \\ 
 & 25\% & 67.81 & 63.41 & \textbf{37.56} & \textbf{76.00} & 86.71 & 62.58 & 46.30 & 79 & \textbf{2.52}x & 30.94 & \textbf{1.24}x & 26.54  \\ 
\bottomrule
\end{tabular}
\caption{Accuracy performance and inference efficiency under different LVLMs and ratios on FoPru. Inference efficiency is measured on the POPE dataset. Results better compared to no pruning are \textbf{bold}. \textit{ms} denotes milliseconds, $S_p$ represents the speedup ratio, \textit{ms/tok.} indicates milliseconds per token.}
\vspace{-5mm}
\label{tab:model_performance}
\end{table*}

\subsection{Experimental Setting}
\paragraph{\textit{Datasets}}
We utilize 7 widely used multimodal datasets to evaluate the performance, including POPE \cite{li2023evaluating}, MMMU \cite{yue2024mmmu}, SQA \cite{lu2022learn}, Ai2D \cite{kembhavi2016diagram}, GQA \cite{hudson2019gqa}, TextVQA \cite{singh2019towards} and Ocrbench \cite{liu2023hidden}. 
POPE is utilized to evaluate the model's ability to identify and correct errors in images within multimodal scenarios. 
MMMU is a multimodal benchmark that covers multiple academic tasks, requiring university-level subject knowledge and reasoning skills. 
SQA (i.e., ScienceQA) focuses on answering questions in the science domain, covering a wide range of topics from basic science to advanced research. 
Ai2D is used to evaluate the model's ability to interpret and understand complex scientific and educational diagrams. 
GQA focuses on visual question answering tasks, testing the model's ability to understand and answer questions about image content.
TextVQA involves processing and answering questions related to text found in images, requiring the model to recognize and understand the text within the images. 
Ocrbench concentrates on optical character recognition (OCR), evaluating the model's ability to recognize and interpret text in images of varying quality and text.

\paragraph{\textit{Implementation Details}} To ensure fairness, all experiments were conducted on the LMMS-Eval platform \cite{zhang2024lmms}, a unified evaluation framework for multimodal models that supports over 50 datasets and more than 10 LVLMs, and is designed to provide transparent and reproducible evaluations.
We selected \texttt{LLaVA-NeXT-8B}~\cite{li2024llavanext-strong}, \texttt{LLaVA-1.6-7B} and \texttt{LLaVA-1.6-13B}~\cite{liu2024llavanext} for our experiments. We implemented our FoPru method with aforementioned token pruning strategies and three different retention proportions of visual tokens, including 25\%, 50\%, and 75\%. We also used Time To First Token (TTFT), which measures the latency from input to the output of the first token, Time Per Output Token (TPOT), which measures the latency per output token, and GPU usage to evaluate inference efficiency.

\subsection{Main Results}

\begin{figure}[h]
\centering
\includegraphics[width=\columnwidth]{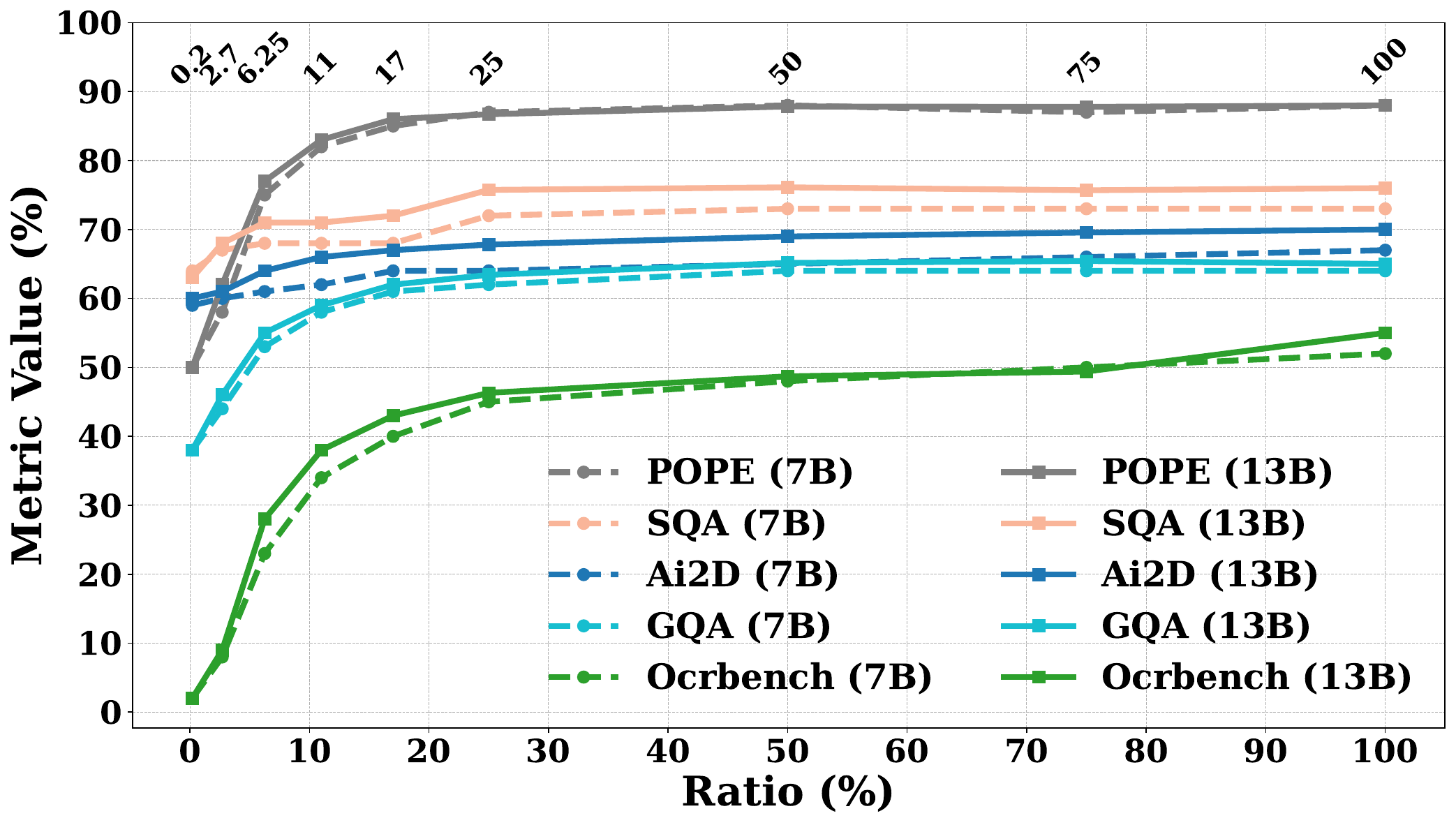}
\caption{Performance metrics across visual token retention ratios for the \texttt{LLaVA-1.6-7B} and \texttt{LLaVA-1.6-13B} models on five datasets.}
\label{fig:ratios}
\vspace{-5mm}
\end{figure}

In this study, we evaluate the performance of FoPru under different token retention ratios. Results are presented in two parts: Figure \ref{fig:ratios} illustrates accuracy trends across different retention ratios to reveal overall patterns, while Table \ref{tab:model_performance} provides a detailed breakdown of accuracy performance and inference efficiency, including the results of FoPru’s best-performing strategy. Comparisons of alternative strategies will be discussed in subsequent sections.

\paragraph{\textit{Token Retention Ratios in Rank Pruning}} Figure~\ref{fig:ratios} illustrates the accuracy performance across nine token retention ratios (0.2\%, 2.7\%, 6.25\%, 11\%, 17\%, 25\%, 50\%, 75\%, and 100\%) for two models across five datasets.
Model accuracy generally improves as the retention ratio increases, but the retention threshold at which accuracy stabilizes varies across datasets. 
SQA, Ai2D, and POPE reach over 50\% accuracy with just 11\% retention, while OCRbench is highly sensitive to retention changes, with performance improving continuously across higher ratios. Specifically, the OCRbench dataset's accuracy rises sharply from 2\% at 0.2\% retention to over 50\% at full retention, suggesting its reliance on a larger proportion of visual tokens. 
Surprisingly, Ai2D and SQA maintain relatively high accuracy (around 60\%) even at extreme levels of token pruning (0.2\% retention, approximately 5 visual tokens or fewer). This robustness at low retention suggests that these datasets may involve tasks where essential information is concentrated within a few core tokens, possibly because the tasks are less visually complex or involve less semantic information in the images. 
These findings indicate that while higher token retention ratios generally enhance model accuracy, the optimal retention levels are dataset-specific, indicating that a flexible, task-oriented approach to token retention 
may benefit LVLMs.

\begin{table*}[ht]
\centering
\setlength\tabcolsep{2.6 pt}
\renewcommand{\arraystretch}{0.8}
\begin{tabular}{cc|ccccccc|ccccc} 
\toprule
\multirow{3}{*}{\textbf{Ratio}} & \multirow{3}{*}{\textbf{Method}} & \multicolumn{7}{c|}{\textbf{Accuracy Performance}} & \multicolumn{5}{c}{\textbf{Inference Efficiency}} \\ 
 &  & \multirow{2}{*}{\textbf{Ai2D}} & \multirow{2}{*}{\textbf{GQA}} & \multirow{2}{*}{\textbf{MMMU}} & \multirow{2}{*}{\textbf{SQA}} & \multirow{2}{*}{\textbf{POPE}} 
 & \multirow{2}{*}{\textbf{TextVQA}}
 & \multirow{2}{*}{\textbf{Ocrbench}} & \multicolumn{2}{c}{\textbf{TTFT}} & \multicolumn{2}{c}{\textbf{TPOT}} & \textbf{GPU} \\ 
 &  &  &  &  &  &  &  &  & (ms & $S_p$) & (ms/tok. & $S_p$) & GB \\ 
\midrule
100\% & - & 55.50 & 61.97 & 35.30 & 69.51 & 86.98 & 46.00 & 31.20 & 74 & - & 27.71 & - & 13.91 \\ 
\midrule
\multirow{2}{*}{Dynamic} 
& PurMerge &52.49	&51.48	&Failed &68.96	&74.61	&37.64	&25.60	 &47	&1.59x	&21.95	&1.26x	&\textbf{13.35}\\
& PruMerge+ & 54.14 & 57.35 & Failed & \textbf{69.16} & 84.25 & 39.41 & 28.00 & 74 & 1.00x & 21.82 & 1.27x & 13.46 \\ 
\midrule
\multirow{2}{*}{75\%} & FastV & 55.34 & \textbf{61.61} & \underline{36.11} & \textbf{69.51} & 86.69 & \underline{46.08} & \underline{31.30} & 69 & 1.07x & 27.40 & 1.01x & 13.81 \\ 
 & \textbf{FoPru} & \textbf{\underline{55.70}} & 61.55 & \textbf{\underline{36.67}} & 68.52 & \textbf{86.92} & 45.87 & \textbf{\underline{31.70}} & \textbf{62} & \textbf{1.20x} & \textbf{21.45} & \textbf{1.29x} & \textbf{13.76} \\ 
\midrule
\multirow{2}{*}{50\%} & FastV & 55.08 & 60.33 & \underline{35.89} & \textbf{68.67} & 85.20 & 45.51 & 30.60 & 59 & 1.26x & 27.16 & 1.02x & 13.69 \\ 
 & \textbf{FoPru} & \textbf{55.12} & \textbf{60.98} & \textbf{\underline{36.11}} & 68.07 & \textbf{\underline{87.40}} & \textbf{45.64} & \textbf{\underline{31.30}} & \textbf{49} & \textbf{1.52x} & \textbf{21.41} & \textbf{1.29x} & \textbf{13.61} \\ 
\midrule
\multirow{3}{*}{25\%} 
 & FastV & 53.95 & 57.47 & \underline{35.44} & 68.86 & 81.21 & 42.56 & 29.00 & 51 & 1.44x & 27.28 & 1.02x & 13.52 \\ 
 & \textbf{FoPru}+FastV & 54.18 & \textbf{58.61} & \underline{36.22} & 68.42 & 84.44 & 44.43 & 30.20 & 49 & 1.50x & 27.12 & 1.02x & 13.51 \\ 
 & \textbf{FoPru} & \textbf{54.60} & 57.94 & \textbf{\underline{36.56}} & 68.27 & \textbf{85.07} & \textbf{44.35} & \textbf{30.50} & \textbf{38} & \textbf{1.97x} & \textbf{20.89} & \textbf{1.33x} & 13.51 \\ 
\bottomrule
\end{tabular}
\caption{Accuracy and inference efficiency of different methods using \texttt{LLaVA-1.5-7B}. Inference efficiency is measured on the POPE dataset. FoPru+FastV indicates FoPru and FastV each prune half of tokens, resulting in 25\% token retention. The best results are \textbf{bold}. Accuracy results that are better compared to no pruning are \underline{underlined}. \textit{ms} denotes milliseconds, $S_p$ represents the speedup ratio, \textit{ms/tok.} indicates milliseconds per token.
\textit{Failed} entries indicate cases where the algorithm could not process the input due to its limitation of not supporting multi-image input.
}
\vspace{-5mm}
\label{tab:compare_method}
\end{table*}

\paragraph{\textit{Accuracy Performance}} As shown in Table \ref{tab:model_performance} \textit{Accuracy Performance}, overall, FoPru can effectively prune a significant number of redundant tokens without sacrificing much accuracy, enabling efficient inference.
Specifically, we have the following detailed findings:
\begin{itemize}
    \item With 25\% token retention, the three LVLMs exhibit less than 1\% accuracy loss on the MMMU, SQA, and POPE datasets. This result supports our hypothesis regarding the redundancy of visual tokens and demonstrates that our method can effectively select the visual tokens containing core information.
    \item With 50\% token retention, the accuracy loss is only around 1\% on the Ai2D, GQA, MMMU, SQA, and POPE datasets. This indicates that for these discriminative VQA tasks, it is unnecessary to include all detailed visual tokens and a small portion of tokens that contain holistic information is sufficient to accomplish them.
    \item On the TextVQA and Ocrbench datasets, pruning tokens results in a relatively large drop in accuracy. We believe this is because the images in these datasets contain much text, which requires the capture of more continuous information from the images.
    \item Despite reducing the number of tokens, FoPru achieves performance surpassing the baseline of retaining all tokens on the MMMU, SQA, POPE, and GQA datasets. This suggests that redundant visual tokens might even interfere with the LVLMs' ability to make accurate judgments. Identifying and removing unimportant tokens can not only accelerate inference but also have the potential to further enhance performance.

\end{itemize}

\paragraph{\textit{Inference Efficiency}}
Since the inference performance of several strategies is close, we only present the results for the rank strategy here, with additional details provided in the Appendix. As shown in Table~\ref{tab:model_performance} \textit{Inference Efficience}, evaluated on the POPE dataset, our results demonstrate that the FoPru method consistently enhances inference efficiency across different LVLMs and retention ratios. 
For the largest model, \texttt{LLaVA-1.6-13B}, the TTFT achieves a speedup of up to \textbf{2.52x}. Similarly, TPOT results in a \textbf{1.24x} speedup. 
Notable variations in GPU memory savings between models can be attributed to architectural differences, where larger models benefit more significantly from token pruning. 
Overall, the results suggest that FoPru can consistently reduce inference time while also lowering GPU memory usage across three LVLMs and ratios, which shows a significant improvement of FoPru in both inference speed and resource consumption.

\subsection{Comparative Experiments}

To further validate the effectiveness of FoPru, 
we conduct experiments on \texttt{LLaVA-1.5-7B}~\cite{liu2024llava15}, comparing it with three recent training-free token pruning methods for LVLMs: FastV, PruMerge, and PruMerge+. 
For FastV, we set token pruning to occur at the third layer of the LLMs, while PruMerge and PruMerge+ dynamically select the most crucial visual tokens to retain without direct control over the retention ratio, with an average retention rate of approximately 5.5\% and 25.0\%, respectively \cite{shang2024llava}. 
All experiments are re-evaluated on the LMMS-Eval platform to ensure consistency and fairness.
Additionally, we also combine FoPru with FastV for further comparison. 
Specifically, FoPru is used to prune half of the tokens before the projector, followed by FastV to prune half of the remaining tokens within the LLM, resulting in 25\% token retention. The results are presented in Table \ref{tab:compare_method}.

As shown in Table \ref{tab:compare_method} \textit{Accuracy Performance}, with 25\% and 50\% token retention, the accuracy of FoPru surpasses that of FastV across six datasets. 
This suggests that FoPru, which leverages the visual encoder’s attention map to prune tokens before they are input into the projector, demonstrates superiority in early removal of redundant tokens compared to FastV’s token pruning within the later LLM. 
As illustrated in Table \ref{tab:compare_method} \textit{Inference Efficiency}, FoPru consistently outperforms FastV in both inference speed and GPU usage across all datasets.
Compared to the adaptive methods PruMerge and PruMerge+, FoPru consistently achieves higher accuracy and inference efficiency, especially on GQA and TextVQA.
Moreover, FoPru+FastV achieves accuracy that surpasses FastV at 25\% token retention across six datasets but remains lower than FoPru alone at the same retention level, emphasizing FoPru’s advantage in early-stage pruning while suggesting potential for more effective integration strategies.

\subsection{Ablation Studies}
\paragraph{\textit{Comparisons of Different Strategies}}
\begin{figure*}[t!]
\centering
\includegraphics[width=\textwidth]{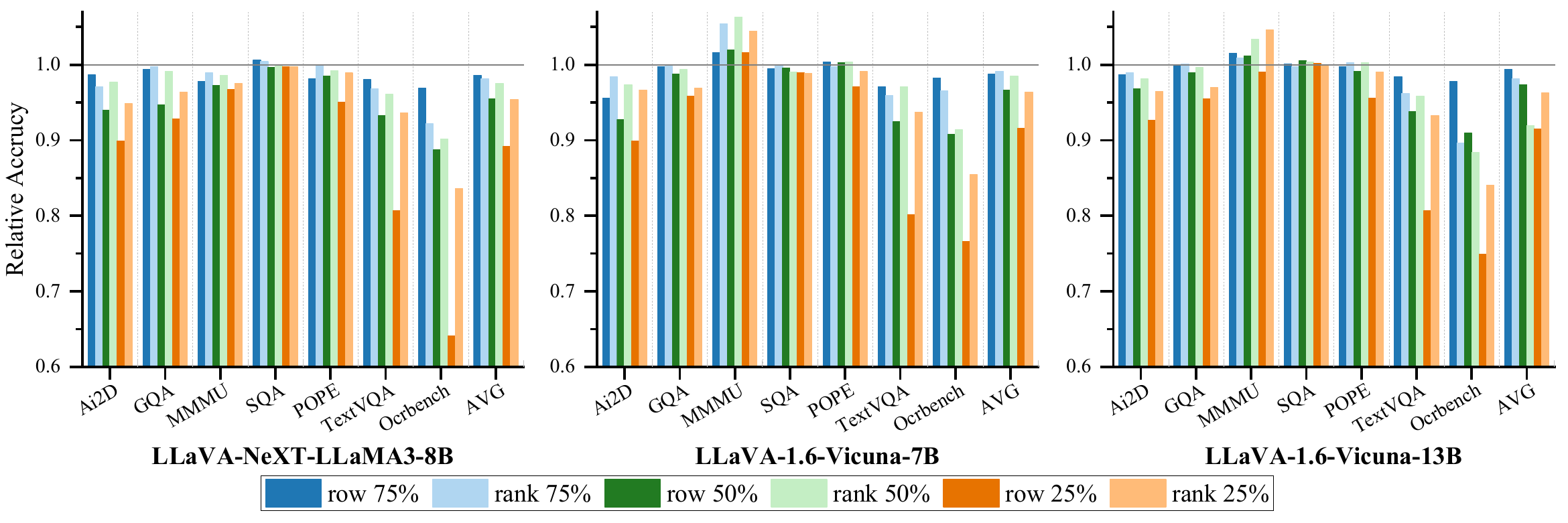} 
\caption{The comparison of different token pruning strategies across different LVLMs and datasets. \textit{Relative Accuracy} is defined as the accuracy accuracy after pruning divided by the accuracy without pruning.
\textit{AVG} denotes the average accuracy across all datasets.}
\label{fig:compare_stratages}
\vspace{-3mm}
\end{figure*}

This study evaluates row and rank pruning strategies across three models on seven datasets, with pruning ratios of 25\%, 50\%, and 75\%. 
To highlight differences, we use \textit{Relative Accuracy}—accuracy after pruning divided by accuracy without pruning, with absolute accuracy results in the appendix.
As shown in Figure \ref{fig:compare_stratages}, the results reveal distinct patterns depending on the pruning ratio: at lower retention ratios, rank pruning generally provides more stable performance across datasets, while at higher ratios, the two strategies converge, with row pruning showing a slight edge. 
Sensitivity to pruning, meanwhile, varies significantly across datasets. For example, datasets such as SQA, POPE, and GQA display minimal accuracy loss—and even occasional improvements
at higher pruning levels, indicating a resilience to token reduction. 
Conversely, OCRbench and TextVQA exhibit marked accuracy improvements as token retention increases with row pruning, suggesting that continuity of token information is essential for tasks requiring detailed visual comprehension. 
These findings highlight the need for task-specific pruning configurations: row pruning is more effective at higher retention levels, especially for visual tasks, while rank pruning maintains strong performance even at lower ratios by capturing global information efficiently. Overall, the results emphasize the importance of tailoring pruning strategies and ratios to optimize performance across tasks.

\paragraph{\textit{Token Significance and Positional Sensitivity}}
\begin{table}[t]
\centering
\renewcommand{\arraystretch}{0.8}
\begin{tabular}{ccccc} 
\toprule
\textbf{Method}  & \textbf{Ai2D}  & \textbf{SQA}   & \textbf{POPE}  & \textbf{TextVQA}  \\ 
\midrule
-             & 71.66          & 79.44          & 87.84          & 65.43             \\ 
\midrule
w/o Variance &67.49 &78.14 &80.54 &51.08 \\
w/o Significance & 67.81          & \textbf{79.51} & 84.42          & 48.38             \\
w/o Reordering       & 67.00          & 78.59          & 86.51          & 53.18             \\
FoPru               & \textbf{68.01} & 79.27          & \textbf{86.88} & \textbf{61.24}    \\

\bottomrule
\end{tabular}
\caption{Ablation study of FoPru on accuracy across multiple datasets using \texttt{LLaVA-NeXT-8B} with 25\% token retention. 
\textit{w/o Variance} refers to selecting low-variance directions for token significance.
\textit{w/o Significance} means replacing the Token Significance stage with a pooling operation (i.e., merging every four adjacent tokens into one). \textit{w/o Reordering} means removing the Token Reordering stage.
The best results are \textbf{bold}.}
\label{tab:ablation}
\vspace{-5mm}
\end{table}
An ablation study was conducted to assess the impact of specific procedural elements within our methodology. 
For comparative analysis, we used the rank strategy retaining 25\% of the tokens. As shown in Table~\ref{tab:ablation}, 
without variance-based token significance (w/o Variance), we observe a notable drop in performance across all datasets, underscoring the importance of high-variance features for capturing salient information. 
Similarly, excluding the Token Reorder phase (w/o Reordering) also leads to consistent performance declines, particularly in Ai2D and TextVQA, indicating that maintaining positional integrity is crucial in tasks with high spatial sensitivity. 
Moreover, removing the significance stage entirely (w/o Significance) by merging adjacent tokens leads to marked performance declines, particularly on POPE and TextVQA, highlighting the value of attention-based token selection. 
Interestingly, the results for SQA increase slightly in the absence of the Token Significance stage. This outcome suggests that SQA may be less reliant on the spatial token arrangement, potentially benefiting from a more condensed representation.

\label{sec:experiments}
\section{Conclusion and Limitations}
\label{sec:conclusion}
This paper proposes a training-free and general inference optimization method for various LVLMs, called Focal Pruning (FoPru),
which aims to address the issue of inefficient inference caused by a large amount of redundancy in visual tokens. Specifically, FoPru leverages the attention scores from the visual encoder to determine the significance score of visual tokens. 
Then, we provide multiple alternative attention-based token pruning strategies before the tokens are input into the projector to significantly reduce the number of visual tokens that the LLM needs to process. 
Finally, a token reordering method is employed to ensure that the relative positional information between tokens is preserved. 
Extensive experiments on three widely used LVLMs and seven multimodal datasets demonstrate that our FoPru can significantly reduce the number of visual tokens and improve inference speed while keeping the accuracy loss minimal. 
While  FoPru demonstrates strong performance on general LVLMs, there are still some limitations. 
For example, the optimal pruning ratios for visual tokens vary across different tasks and models, and they remain unclear. 
Additionally, we observe that FoPru requires retaining a higher number of visual tokens for tasks that rely heavily on positional information. 
In the future, we will explore combining FoPru with more existing inference optimization techniques for LVLMs to further boost inference efficiency.

{
    \small
    \bibliographystyle{ieeenat_fullname}
    \bibliography{main}
}

\clearpage
\setcounter{page}{1}

\section{Appendix}
\begin{figure}[]
\centering
\includegraphics[width=\columnwidth]{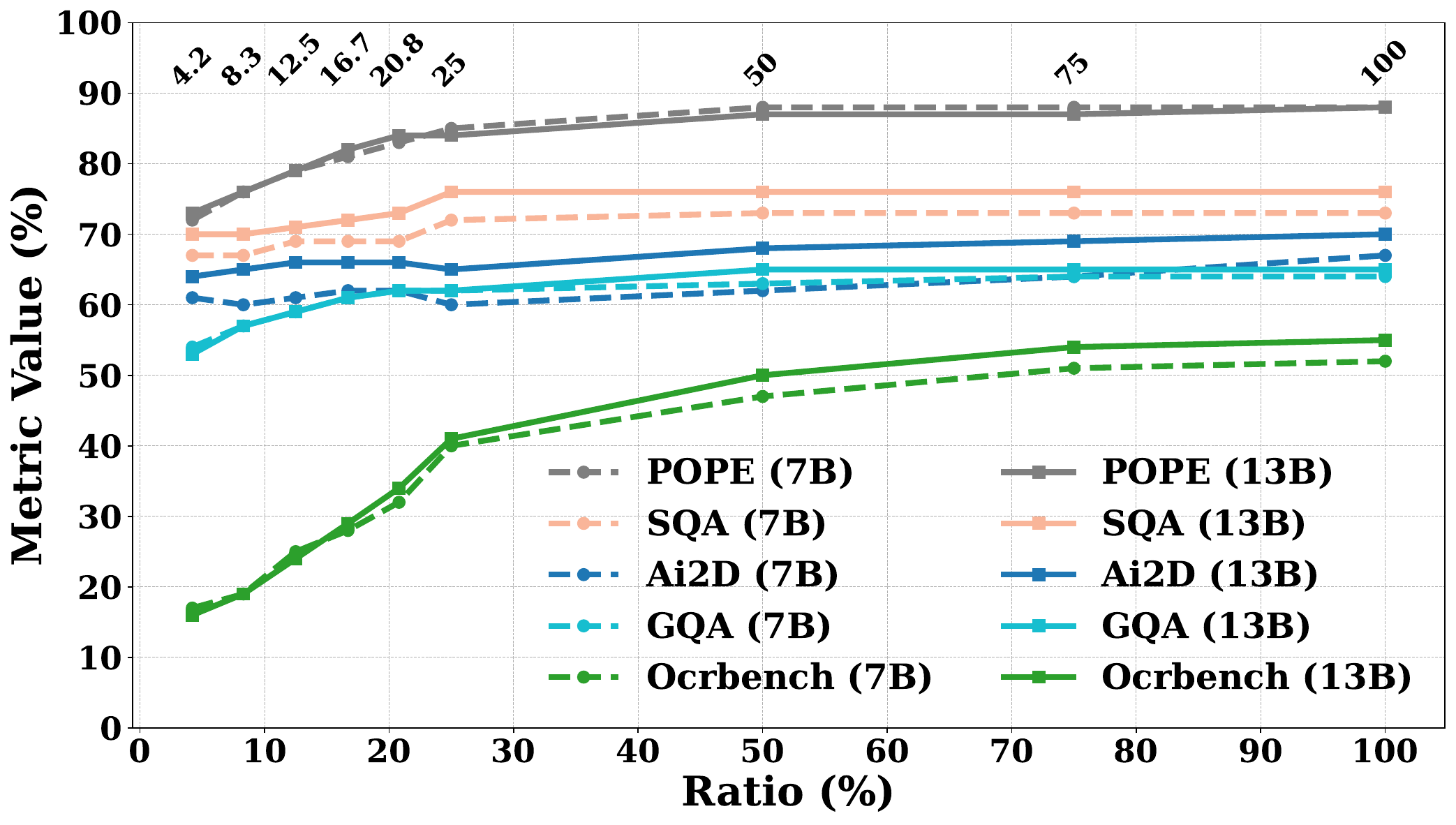}
\caption{Performance metrics across visual token retention ratios for the \texttt{LLaVA-1.6-7B} and \texttt{LLaVA-1.6-13B} models on five datasets.}
\label{fig:row_ratios}
\end{figure}

\paragraph{\textit{Token Retention Ratios in Row Pruning}} 
Figure~\ref{fig:row_ratios} shows that accuracy increases steadily with higher token retention ratios for both models under row pruning. 
The trend is smoother compared to rank pruning, where accuracy often rises more sharply at lower ratios (e.g., Ocrbench).
Both strategies converge at higher retention levels (50\%-100\%), achieving similar performance. 
Overall, rank pruning shows faster gains at lower ratios, while row pruning provides more consistent improvements across all ratios.

\paragraph{\textit{Accuracy Performance}}
We conducted a comprehensive performance evaluation of various models using different pruning ratios and strategies across multiple tasks. The results are summarized in the Table~\ref{tab:model_ratio_strategy_comparison}.
\begin{table}[]
\centering
\renewcommand{\arraystretch}{0.4}
\resizebox{0.48\textwidth}{!}{
\setlength{\tabcolsep}{1pt} %
\begin{tabular}{lllcccccccc}
\toprule
\textbf{Model} & \textbf{Ratio} & \textbf{Strategy} & \textbf{Ai2D} & \textbf{TextVQA} & \textbf{GQA} & \textbf{MMMU} & \textbf{SQA} & \textbf{POPE} & \textbf{Ocrbench} & \textbf{Avg} \\
\midrule

\multirow{7}{*}{\begin{tabular}[c]{@{}c@{}}\\ \\ \\ \\LLaVA-\\NeXT-\\8B\end{tabular}}
& base & -    & 71.66  & 65.43  & 65.38  & 40.22  & 79.44  & 87.84  & 54.90  & 66.41  \\
\cmidrule{2-11}
& \multirow{2}{*}{75\%}  
& row  & \underline{70.69}  & \underline{64.14}  & 64.96  & 39.33  & \textbf{\underline{79.91}}  & 86.19  & \underline{53.20}  & \underline{65.49}  \\
&   
& rank & 69.56  & 63.35  & \underline{65.21}  & \underline{39.78}  & 79.77  & \textbf{\underline{87.87}}  & 50.60  & 65.16  \\
\cmidrule{2-11}
& \multirow{2}{*}{50\%}  
& row  & 67.33  & 61.02  & 61.89  & 39.11  & 79.16  & 86.51  & 48.70  & 63.39  \\
&   
& rank & \underline{70.02}  & \underline{62.86}  & \underline{64.82}  & \underline{39.67}  & \underline{79.39}  & \underline{87.13}  & \underline{49.50}  & \underline{64.77}  \\
\cmidrule{2-11}
& \multirow{2}{*}{25\%}  
& row  & 64.44  & 52.79  & 60.68  & 38.89  & 79.25  & 83.50  & 35.20  & 59.25  \\
&   
& rank & \underline{68.01}  & \underline{61.24}  & \underline{63.00}  & \underline{39.22}  & \underline{79.27}  & \underline{86.88}  & \underline{45.90}  & \underline{63.36}  \\
\midrule

\multirow{7}{*}{\begin{tabular}[c]{@{}c@{}}\\ \\ \\ \\ \\LLaVA-\\1.6-7B\end{tabular}}
& base & -    & 66.58  & 64.90  & 64.24  & 35.10  & 73.21  & 87.61  & 52.20  & 63.41  \\
\cmidrule{2-11}
& \multirow{2}{*}{75\%}  
& row  & 63.63  & \underline{63.00}  & 64.07  & \textbf{35.67}  & 72.86  & \textbf{\underline{87.93}}  & \underline{51.30}  & 62.64  \\
&   
& rank & \underline{65.54}  & 62.25  & \underline{64.13}  &\textbf{\underline{37.00}}  & \underline{73.19}  & 87.39  & 50.40  & \underline{62.84}  \\
\cmidrule{2-11}
& \multirow{2}{*}{50\%}  
& row  & 61.72  & 60.00  & 63.46  & \textbf{35.78}  & \underline{72.91}  & \textbf{87.86}  & 47.40  & 61.30  \\
&   
& rank & \underline{64.83}  & \underline{63.01}  & \underline{63.83}  & \textbf{\underline{37.33}}  & 72.53  & \textbf{\underline{87.93}}  & \underline{47.70}  & \underline{62.45}  \\
\cmidrule{2-11}
& \multirow{2}{*}{25\%}  
& row  & 59.84  & 52.03  & 61.57  & \textbf{35.67}  & \underline{72.41}  & 85.02  & 40.00  & 58.08  \\
&   
& rank & \underline{64.35}  & \underline{60.81}  & \underline{62.26}  & \textbf{\underline{36.67}}  & 72.39  & \underline{86.83}  & \underline{44.60}  & \underline{61.13}  \\
\midrule

\multirow{7}{*}{\begin{tabular}[c]{@{}c@{}}\\ \\ \\ \\ \\ LLaVA-\\1.6-13B\end{tabular}}
& base & -    & 70.30  & 67.10  & 65.37  & 35.90  & 75.85  & 87.56  & 55.10  & 65.31  \\
\cmidrule{2-11}
& \multirow{2}{*}{75\%}  
& row  & 69.40  & \underline{66.03}  & 65.37  & \textbf{\underline{36.44}}  & \textbf{\underline{75.95}}  & 87.33  & \underline{53.90}  & \underline{64.92}  \\
&   
& rank & \underline{69.56}  & 64.54  & \textbf{\underline{65.43}}  & \textbf{36.22}  & 75.69  & \textbf{\underline{87.78}}  & 49.40  & 64.09  \\
\cmidrule{2-11}
& \multirow{2}{*}{50\%}  
& row  & 68.07  & 62.95  & 64.68  & \textbf{36.33}  & \textbf{\underline{76.23}}  & 86.81  & \underline{50.10}  & \underline{63.60}  \\
&   
& rank & \underline{68.98}  & \underline{64.33}  & \underline{65.15}  & \textbf{\underline{37.11}}  & \textbf{76.11}  & \textbf{\underline{87.84}}  & 48.70  & 60.06  \\
\cmidrule{2-11}
& \multirow{2}{*}{25\%}  
& row  & 65.12  & 54.16  & 62.45  & 35.56  & \textbf{\underline{76.00}}  & 83.67  & 41.30  & 59.75  \\
&   
& rank & \underline{67.81}  & \underline{62.58}  & \underline{63.41}  & \textbf{\underline{37.56}}  & 75.74  & \underline{86.71}  & \underline{46.30}  & \underline{62.87}  \\

\bottomrule
\end{tabular}
}
\caption{Accuracy performance comparison across different models, ratios, and strategies. \textit{Avg} refers to the average accuracy across tasks.
The best results are \underline{underlined}. 
Accuracy results that are better compared to not pruning are \textbf{bold}.}
\label{tab:model_ratio_strategy_comparison}
\end{table}

\paragraph{\textit{Inference Efficiency}}
Table~\ref{tab:row} shows the inference efficiency using $\textit{row strategy}$ under different LVLMs and ratios on the POPE dataset.
The results demonstrate inference speedups similar to those observed with the $\textit{rank pruning}$ strategy.

\begin{wraptable}{l}{\columnwidth} 
\centering
\setlength\tabcolsep{1.5pt} 
\resizebox{0.4\textwidth}{!}{

\begin{tabular}{ccccccc} 
\toprule
\multirow{2}{*}{\textbf{Model}}                                          & \multirow{2}{*}{\textbf{Ratio}} & \multicolumn{2}{c}{\textbf{TTFT}} & \multicolumn{2}{c}{\textbf{TPOT}} & \textbf{GPU}  \\ 
\cmidrule(l){3-7}
&  & ms  & $S_p$    & ms/tok. & $S_p$    & GB                         \\ 
\midrule
\multirow{4}{*}{\begin{tabular}[c]{@{}c@{}}LLaVA-\\NeXT-8B\end{tabular}} & 100\%                            & 94 & -                           & 25.61   & -                       & 17.88                           \\ 
\cmidrule(l){2-7}
        ~ & 75\% & 74 & 1.28x & 24.72 & 1.04x & 17.28  \\ 
        ~ & 50\% & 62 & 1.52x & 24.57 & 1.04x & 16.98  \\ 
        ~ & 25\% & 53 & 1.78x & 23.67 & 1.08x & 16.98  \\ 
        \midrule
\multirow{4}{*}{\begin{tabular}[c]{@{}c@{}}LLaVA-\\1.6-7B\end{tabular}} & 100\% & 88 & - & 23.70 & - & 16.15  \\ 
        ~ & 75\% & 76 & 1.17x & 23.48 & 1.01x & 15.40  \\ 
        ~ & 50\% & 53 & 1.67x & 23.10 & 1.03x & 14.75  \\ 
        ~ & 25\% & 52 & 1.71x & 22.82 & 1.04x & 14.57  \\       
\midrule 
       
\multirow{4}{*}{\begin{tabular}[c]{@{}c@{}}LLaVA-\\1.6-13B\end{tabular}} & 100\% & 198 & - & 38.30 & - & 29.53  \\ 
        ~ & 75\% & 153 & 1.29x & 35.66 & 1.07x & 28.44  \\ 
        ~ & 50\% & 110 & 1.80x & 33.16 & 1.15x & 27.48  \\ 
        ~ & 25\% & 76 & 2.61x & 30.84 & 1.24x & 26.42  \\ 
\bottomrule       
\end{tabular}
}
\caption{Inference efficiency using \textit{row\ pruning} strategy under different LVLMs and ratios on the POPE dataset. \textit{ms} denotes milliseconds, $S_p$ represents the speedup ratio, \textit{ms/tok.} indicates milliseconds per token.}
\label{tab:row}
\end{wraptable}

\end{document}